%% file: main.tex
\documentclass[runningheads]{llncs}
\usepackage{graphicx}
\usepackage{amsmath,amssymb}
\usepackage{color}
\usepackage{subfigure}
\usepackage{multirow}
\usepackage{tabularx}
\usepackage{booktabs}
\usepackage{wrapfig}
\usepackage{comment}
\usepackage{makecell}
\usepackage{hyperref}
\input{defs}

\begin{document}
\title{DIG: Draping Implicit Garment over the Human Body\thanks{This work was supported in part by the Swiss National Science Foundation.}}
%
\titlerunning{DIG: Draping Implicit Garment over the Human Body}
%
\author{Ren Li\inst{1}\orcidID{0000-0003-2998-7104} \and
Benoît Guillard\inst{1}\orcidID{0000-0002-8747-6153} \and
Edoardo Remelli\inst{2}\orcidID{0000-0002-8506-9191} \and
Pascal Fua\inst{1}\orcidID{0000-0002-5477-1017}}
%
\authorrunning{R. Li et al.}
%
\institute{CVLab, EPFL, Switzerland \\
\email{\{ren.li, benoit.guillard, pascal.fua\}@epfl.ch}\\
\and
Meta Reality Labs Research, Zurich, Switzerland\\
\email{edoremelli@fb.com}}
\maketitle 
\input{tex/0_abstract}
\input{tex/1_introduction}
\input{tex/2_related_work}
\input{tex/3_method}
\input{tex/5_experiments}
\input{tex/6_conclusion}
%

%
%
%

%
%
%
%

\bibliographystyle{splncs04}
\bibliography{string,graphics,vision,optim,misc,learning,local}

\end{document}



\title{DIG: Draping Implicit Garment over the Human Body \\--- Supplementary Material ---}
\author{Ren Li\inst{1}\orcidID{0000-0003-2998-7104} \and
Benoît Guillard\inst{1}\orcidID{0000-0002-8747-6153} \and
Edoardo Remelli\inst{2}\orcidID{0000-0002-8506-9191} \and
Pascal Fua\inst{1}\orcidID{0000−-0002-−5477-−1017}}
\titlerunning{DIG - Supplementary Material}
\authorrunning{R. Li et al.}

\institute{CVLab, EPFL, Switzerland \\
\email{\{ren.li, benoit.guillard, pascal.fua\}@epfl.ch}\\
\and
Meta Reality Labs Research, Zurich, Switzerland\\
\email{edoremelli@fb.com}}

\maketitle
%
%
\input{tex/7_supp}

\bibliographystyle{splncs04}
\bibliography{string,graphics,vision,optim,misc,learning,local}

%% file: defs.tex
\newif\ifdraft
\draftfalse
\drafttrue

\usepackage[table]{xcolor}
\definecolor{orange}{rgb}{1,0.5,0}
\definecolor{violet}{RGB}{70,0,170}

\newcommand{\bx}{\mathbf{x}}
\newcommand{\bz}{\mathbf{z}}

\newcommand{\bn}{\mathbf{n}}

%% file: tex/0_abstract.tex
\input{figs/teaser.tex}
\begin{abstract}

Existing data-driven methods for draping garments over human bodies, despite being effective, cannot handle garments of arbitrary topology and are typically not end-to-end differentiable. To address these limitations, we propose an end-to-end differentiable pipeline that represents garments using implicit surfaces and learns a skinning field conditioned on shape and pose parameters of an articulated body model. To limit body-garment interpenetrations and artifacts, we propose an interpenetration-aware pre-processing strategy of training data and a novel training loss that penalizes self-intersections while draping garments. We demonstrate that our method yields more accurate results for garment reconstruction and deformation with respect to state of the art methods. Furthermore, we show that our method, thanks to its end-to-end differentiability, allows to recover body and garments parameters jointly from image observations, something that previous work could not do. Our code is available at \url{https://github.com/liren2515/DIG}.
\end{abstract}

%% file: figs/teaser.tex
\begin{figure}[ht!]
    \centering
    \subfigure[]{\label{fig:teaser1}
    \includegraphics[width=0.35\linewidth]{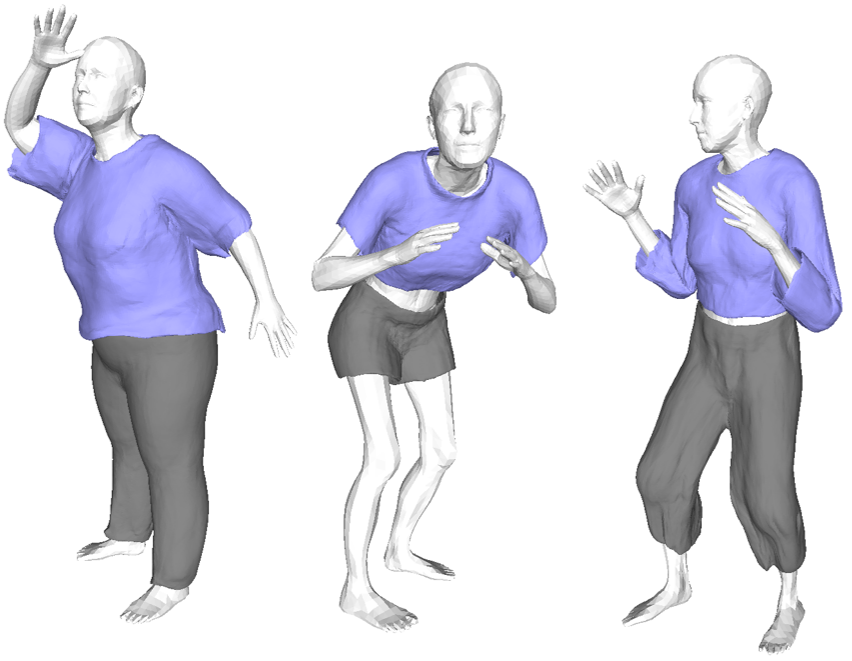}}
    \subfigure[]{\label{fig:teaser2}
    \includegraphics[width=0.6\linewidth]{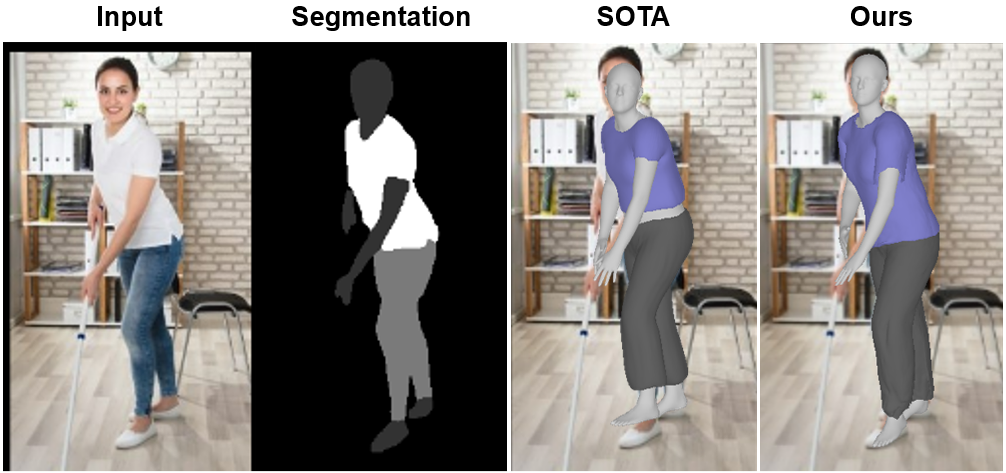}}
    \caption{We introduce a pipeline for (a) generating and draping garments with various topology plausibly and (b) recovering garments from image observations (e.g. segmentation masks). Unlike prior works, our method allows for joint optimization of garment and body meshes, resulting in more faithful reconstruction.}
    \label{fig:teaser}
\end{figure}

%% file: tex/1_introduction.tex
\section{Introduction}

\input{figs/pipeline.tex}

Modeling clothed humans has applications in industries such as fashion design, moviemaking, and video gaming. Many professional tools that rely on Physics-Based Simulation (PBS)~\cite{Nvcloth,Optitext,NvFlex,MarvelousDesigner} can be used to model cloth deformations realistically. However, they are computationally expensive, which precludes real-time use. Some of these can operate in near real-time using an incremental approach in motion sequences. However, these methods remain too slow for static cloth draping over a body in an arbitrary pose. 

In recent years, there has therefore been considerable interest in using data-driven techniques to overcome these difficulties. They fall into two main categories. First there are those that use a single model to jointly represent the person and their clothes~\cite{Ma21,Saito21,Chen21d,Tiwari21,Chen22}.  They produce visually appealing results but, because the body and garment are bound together, they do not make it easy to mix and match different bodies and clothing articles. Second, there are methods  that represent the body and clothes separately. For example, in~\cite{Gundogdu22,Patel20,Santesteban21,Tiwari20,Bertiche21}, deep learning is used to define skinning functions that can be used to deform the garments according to body motion. In~\cite{Corona21}, the explicit representation of clothes is replaced by an implicit one that relies on an inflated SDF surrounding the garment surface. It makes it possible to represent garments with many different topologies using a single model. To this end, it relies on the fact that garments follow the underlying body pose predictably. Hence, for each garment vertex, it uses the blending weights of the closest body vertex in the SMPL model~\cite{Loper14}. Unfortunately, this step involves a search, which makes it both computationally expensive and non-differentiable.

In this paper, we propose the novel data-driven approach to skinning depicted by Fig.~\ref{fig:pipeline}. As in~\cite{Corona21}, we represent the garments in terms of an inflated SDF but, instead of using the SMPL skinning model, we learn a garment-specific one. This makes our approach both more expressive and fully-differentiable. To address the interpenetration issues caused by SDF inflation, we devised an interpenetration-aware data preprocessing for our training data. And to properly regularize the learned skinning field and to prevent self-intersections, we introduce a new loss term whose minimization prevents the creation of garment artifacts when the body deforms.

As a result, our method yields state-of-the-art results for both garment reconstruction and deformation. Its full differentiability makes it possible to fit both body and garments to partial observations.  In other words,  our pipeline can be used to simultaneously optimize the body and garment meshes, whereas earlier work~\cite{Corona21} can only be used to optimize the garment.

%% file: figs/pipeline.tex
\begin{figure}[th!]
    \centering
    \includegraphics[width=1.\linewidth]{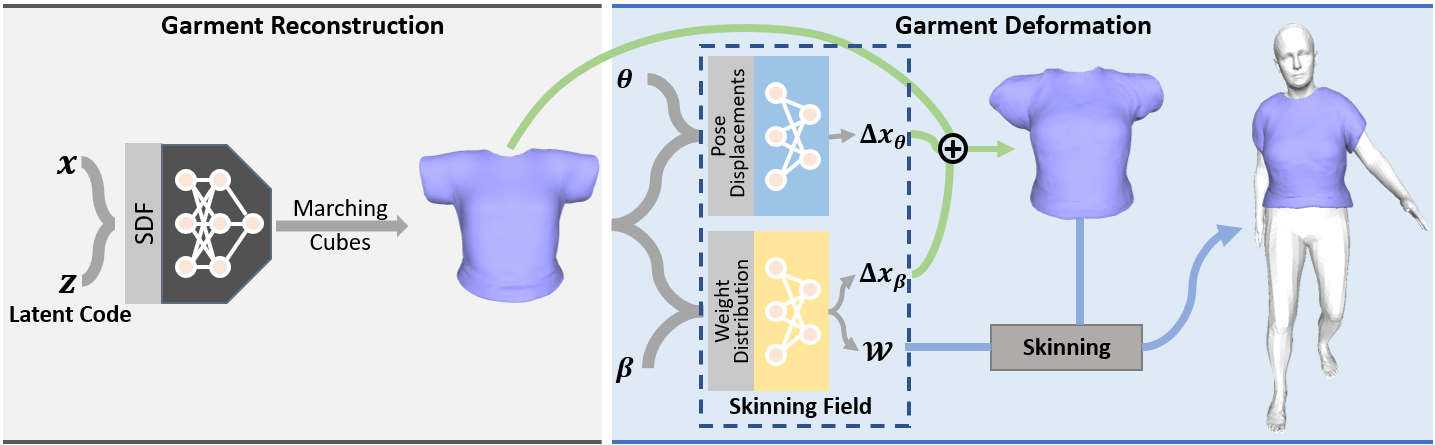}
    \caption{The pipeline of our approach. The garment in the canonical space is first reconstructed from SDF. Given the shape $\beta$ and pose $\theta$ of the target body, we add the shape and pose displacements ($\Delta x_{\beta}$ and $\Delta x_{\theta}$) to the reconstructed garment and drape it to the target body by the skinning function.}
    \label{fig:pipeline}
\end{figure}

%% file: tex/2_related_work.tex

\section{Related Work}

Most garment deformation approaches are either physics-based or data-driven. The physics-based algorithms~\cite{Baraff98,Narain12,Narain13,Liang19a,Li21f} yield highly-realistic deformations but tend to be computationally demanding. The data-driven approaches are much less expensive at inference-time, sometimes at the cost of realism. Here we focus on those that are designed to drape a garment on a posed body. 

\noindent
\textbf{Templates.} In~\cite{Patel20,Santesteban21,Tiwari20,Bhatnagar19,Jiang20d,Bertiche21a,Santesteban22}, individual garments are represented by  separate triangulated 3D meshes. The topology of each one is fixed and a specific deformation function has to be learned. As a result, given the raw scan of a new garment with a different geometry from those already modeled---for example, a skirt as opposed to pants and shorts---expert knowledge is required to create the new template. Furthermore, the deformation model being garment-dependent makes these approaches impractical on large arrays of garments and, hence, ill-suited to real-world applications. 

\noindent
\textbf{Point-Clouds.} In~\cite{Gundogdu22} and DeePSD~\cite{Bertiche21}, the meshes are replaced by clouds of 3D points. The deformation is estimated for each point separately, making it possible to animate outfits of arbitrary topology and geometric complexity. However, the garment topology of these work is still non-differentiable because they rely on vertex connections from the template, which are fixed and pre-designed.  This is addressed in~\cite{Zakharkin21} by using a point-cloud template with a fixed number of points densely sampled from the body mesh. This yields differentiability but the lack of point connections makes the reconstructed garments a group of unordered points instead of a surface with concrete physical properties.

\noindent
\textbf{Implicit Functions.} Deep implicit functions~\cite{Park19c,Chibane20b}  are good at representing surfaces whose topology can change while preserving differentiability~\cite{Remelli20b,Guillard21a}. SMPLicit~\cite{Corona21} is the only work we know of that takes advantage of this to drape garments over bodies. As a result, the model can be fitted  to real-world images. However, SMPLicit suffers several limitations. First, it does not handle the interpenetration between the body and garment. Second, it directly uses the blending weights of the closest vertices in the body model~\cite{Loper14}, which oversimplifies the dynamics and yields over-smoothed results. Finally, the optimization routines used to solve the fitting problem include approximations that produce inaccuracies and prevent the  fitting result from being optimal. Our approach is in a similar spirit but overcomes these limitations. 

%% file: tex/3_method.tex

\section{Method}

We start from an implicit surface model of the garment in a canonical pose, that is, draped over an average body in a T-pose, which we then deform to fit different body shapes and poses. This yields a fully differentiable pipeline that can be used for animation and modeling from images.

\subsection{Garment Representation}
\label{sec:recon}

Watertight surfaces of arbitrary topology can be represented very effectively by the zero crossings of a signed distance function (SDF)
\begin{equation}
    f_{\Theta}(\bx,\bz) \longrightarrow \mathbb{R} \; ,
    \label{eq:sdf}
\end{equation}
where $f$ is implemented by a neural network with weights $\Theta$, $\bx \in \mathbb{R}^3$ is a point in space, and $\bz$ is a latent vector that parameterizes the surface shape~\cite{Park19c}. 
However, clothes have openings in them and are not watertight. To nevertheless represent them in this manner, we can first compute unsigned distances to the surfaces, subtract a small $\epsilon$ value and treat the result as a signed distance function. This amounts to {\it inflating} the garments and representing them as watertight thin surfaces of thickness $2\epsilon$, as in~\cite{Corona21,Guillard22a}. Note that $\epsilon$ cannot be too small and must be larger than marching cube's step size, introducing an undesirable dependency between the field and how it is meshed.

Given a database of garments fitted to a body in a T-pose shape and whose vertices coordinates have been normalized to be between -1 and 1, we use an auto-decoding approach to learning the weights $\Theta$  and the latent vectors $\bz$ associated to specific garments. To this end, for each sample garment and its associated latent vector $\bz$, we minimize a loss function 
\begin{align}
    Loss & = L_{SDF} + \lambda_{grad}L_{grad} + \lambda_{reg}  \lVert \bz \rVert^2  \; \label{eq:sdf-loss} ,\\
    L_{SDF}  &= \sum_{\bx \in X_v} \lVert f_{\Theta}(\bx,\bz)-s^{gt}(\bx) \rVert \; , \label{eq:sdf-loss1} \\
    L_{grad}  &= \sum_{\bx \in X_s}  \lVert \nabla_x f_{\Theta}(\bx,\bz)-\bn^{gt}(\bx) \rVert^2 + \sum_{\bx \notin X_s}  (\lVert \nabla_x f_{\Theta}(\bx,\bz)\rVert-1) ^2 \; ,  \label{eq:sdf-grad}
\end{align}
where $s^{gt}$ and $\bn^{gt}$ are ground-truth values of the signed distance function and normal, $X_v$ and $X_s$ represent points sampled in the $[-1, 1]^3$ volume and the garment surface respectively, and $\lambda_{grad}$ and $\lambda_{reg}$ are scalars that control the influence of the different terms. Minimizing $L_{SDF}$ ensures that the SDF estimated by $f_{\Theta}$ is close to the ground-truth one in the whole volume while minimizing $L_{grad}$ gives additional emphasis to it producing the right normals close to the surface and being a true SDF with unit gradients elsewhere, as in~\cite{Gropp20}. We present an ablation study in the results section that shows that both are necessary to produce smooth and accurate surfaces.

\input{figs/sdf.tex}

One difficulty with this scheme arises from the fact that the garment is usually close to the underlying body mesh and inflating it by $\epsilon$ results in interpenetrations between garment and body,  as shown in Fig.~\ref{fig:inflate1}. Intersections between garments and the human body are problematic because they do not allow to employ the reconstructed meshes for downstream tasks such as e.g. physics simulations. Furthermore, in the experiment section, we show that learning a physically correct representation of garments where there are no interpenetration results in more accurate clothing deformations. To address this, we perform the interpenetration-aware pre-processing illustrated by Fig.~\ref{fig:inflate2} when sampling the surface points in the $X_s$ set of Eq.~\ref{eq:sdf-grad}. Given a garment mesh $G$, we sample a $256 \times 256 \times 256$ grid in $[-1, 1]^3$ to produce a set of points $X$ and compute their signed distance to $G$. We then  run Marching Cubes to recover the watertight mesh $M_{initial}$ as the dashed line of Fig.~\ref{fig:inflate1}. For any vertex of $M_{initial}$ whose signed distance to the body is negative---meaning that it is inside it---we find the closest body vertex $v_c$ and replace its position by $v_c + \mu \textbf{n}_{v_c}$, where $\textbf{n}_{v_c}$ is the surface normal at $v_c$ and $\mu$ is a small positive value, which finally gives us the mesh $M_{clean}$ without interpenetrations depicted by the blue dashed line of Fig.~\ref{fig:inflate2}. In this example, $X_s$ consists of points sampled from $M_{clean}$ located on that dashed line. $X_v$ comprises the points randomly sampled from $[-1, 1]^3$. Their position is not affected but their ground-truth signed distance is computed with respect to $M_{clean}$.

\subsection{Modeling Garment Deformations}
\label{sec:deform}

SMPL is a statistical parametric model that uses Linear Blend Skinning to deform a rigged body template $\textbf{T}\in \mathbb{R}^{N_B\times 3}$ with $N_B$ vertices. Given the parameters of shape $\beta$ and pose $\theta$, SMPL can generate the body mesh $ M_B(\beta, \theta)$ by
\begin{equation}\label{eq:smpl}
    M_B(\beta, \theta)=W(T_B(\beta, \theta), J(\beta), \theta, \mathcal{W}) \; ,
\end{equation}
\begin{equation}
    T_B(\beta, \theta)=\textbf{T} + B_s(\beta) + B_p(\theta) \; ,
\end{equation}
where $W$ is the skinning function with weight $\mathcal{W}\in \mathbb{R}^{N_B\times 24}$ and joint locations $J(\beta)\in \mathbb{R}^{24\times 3}$. $B_s(\beta)\in \mathbb{R}^{N_B\times 3}$ and $B_p(\theta)\in \mathbb{R}^{N_B\times 3}$ are the shape and pose displacements. The SMPLicit algorithm~\cite{Corona21} exploits the fact that the garment follows the pose of the underlying body in a predictable way by using for each garment vertex the blending weights of the closest body vertex. This step involves a search, which makes it both computationally expensive and non-differentiable.

To remedy this, we instead learn a specific blending model for the garment, which is different from that of the body. More specifically, we write 
\begin{align}
  M_{G}(\bx, \beta, \theta) &= W(\bx_{(\beta,\theta)}, J(\beta), \theta, \mathcal{W}(\bx)) \;  \label{eq:model} , \\
  \bx_{(\beta, \theta)} &= \bx + \Delta x_{\beta}(\bx) + \Delta x_{\theta}(\bx) \; \label{eq:model_x} , \nonumber
\end{align}
where $W(\cdot)$ is the SMPL skinning function with learned skinning weights $\mathcal{W}(x)\in \mathbb{R}^{24}$, $\Delta x_{\beta}(\bx)$ and $\Delta x_{\theta}(\bx)$ are shape and pose displacements, and   $\bx\in \mathbb{R}^3$ denotes a generic 3D point instead of on a template. 
$\Delta x_{\beta}(\bx)$ models the shape offset conditioned on body shape $\beta$, while $\Delta x_{\theta}(\bx)$ represents a deformation field conditioned on body pose $\theta$.

More specifically, $\mathcal{W}(\bx)$ and $\Delta x_{\beta}(\bx)$ are computed using the skinning weight $\mathcal{W}$ and shape displacement $B_s(\beta)$ from SMPL as base priors. They are extended to the whole 3D volume by writing
\begin{equation}\label{eq:ours-weight}
  \mathcal{W}(\bx) = w(\bx)\mathcal{W}, ~~\Delta x_{\beta}(\bx)=w(\bx)B_s(\beta),
\end{equation}
where $w(\bx)\in \mathbb{R}^{N_B}$ are shared weights. Since $w(\cdot)$ is implemented by a neural network and $W(\cdot)$ is a differentiable function, $M_{G}$ is fully differentiable, unlike the SMPLicit model~\cite{Corona21}. The approach of~\cite{Santesteban21} does something similar but in a more complex manner because it needs to learn separate models for blending weights and shape displacement, whereas we need only one. Furthermore, because $\bx$ can be {\it any} 3D point, we can deform garments of arbitrary topology, instead of being restricted to a single garment template as in \cite{Patel20,Jiang20d,Santesteban21}.

\subsection{Training the Model}
\label{sec:training}

To train the network that implements the function $w$ of Eq.~\ref{eq:ours-weight}, we use the same sampling strategy as in~\cite{Santesteban21} to collect target $\Bar{w}(x)$ values. For each $\bx\in \mathbb{R}^3$, we sample $N$ points $\mathcal{P}=\{\textbf{p}:\textbf{p}\sim \mathcal{N}(\bx,d)\}$, where $d$ is the distance from $\bx$ to the body. We take $\Bar{w}(\bx)$ to be
\begin{equation}\label{eq:gt-weight}
  \Bar{w}(\bx) = \frac{1}{N} \sum_{\textbf{p}\in \mathcal{P}}w_{bary}(\phi(\textbf{p})),
\end{equation}
where $\phi(\cdot)$ denotes the closest point on the body surface and $w_{bary}(\cdot)$ is  a $N_B$-vector that uses the barycentric coordinate of the closest point as the weight for each body vertex. Since $\Bar{w}(\bx)$ can be regarded as the weight distribution of body vertices, at training time, we introduce the loss
\begin{equation}\label{eq:kl}
L_{KL}=\sum_xKL(w(\bx)||\Bar{w}(\bx)) \; ,
\end{equation}
where $KL$ is the KL-divergence. After the training of $w(x)$, we fix its parameter weights, plug it into our skinning model (Eq. \ref{eq:model}), and then minimize the following loss for the training of $\Delta x_{\theta}$
\begin{equation}\label{eq:pair}
  Loss = \lambda_{deform} L_{deform} + \lambda_{interp} L_{interp} + \lambda_{order} L_{order}.
\end{equation}
where $L_{interp}$ and $L_{order}$ are regularization terms described below and $\lambda_{deform}$, $\lambda_{interp}$, and $\lambda_{order}$ are scalar weights.

\paragraph{\bf Dynamics.} 
To capture detailed dynamics induced by pose changing, we define the deformation loss
\begin{equation}\label{eq:deform}
  L_{deform} = \sum_{\bx\in X_s} |\Bar{x}_d-\hat{x_d}(\bx)|+\sum_{\bx\notin X_s} |\Delta x_{\theta}(\bx)-\Delta x_{\theta}(\bx_c)| \; ,
\end{equation}
where $X_s$ denotes vertices of the ground-truth garment that forms an open surface, $\hat{x}_d$ and $\Bar{x}_d$ are the point deformed according to Eq. \ref{eq:model} and the corresponding ground-truth position, respectively. $\bx_c=\underset{\bx'\in X_s}{\arg\min} ~d(\bx',\bx) $ denotes the surface point closest to $\bx$. As there are no correspondences in the training data for $\bx\notin X_s$, the second term in Eq.~\ref{eq:deform} allows them be learned under the guidance of the closest surface points in the garment. 

\paragraph{\bf Interpenetrations.} 

To prevent them, we utilize the SDF of the body mesh $M_B(\beta,\theta)$ to penalize the presence of deformed points inside the body. We write
\begin{equation}\label{eq:col}
  L_{interp} = \sum_\bx max(0, \epsilon_{SDF}-SDF_{B}(\hat{x_d}(\bx))) \; ,
\end{equation}
where $\epsilon_{SDF}$ is a small value chosen to prevent $\hat{x_d}(\bx)$ from overlapping with the body surface. 

\paragraph{\bf Self-Intersections.} 

Minimizing $L_{deform}$ and $L_{interp}$ usually suffices to deform open surfaces realistically. Unfortunately, when deforming the inflated watertight meshes we use, self-intersections can appear as shown on the left of Fig.~\ref{fig:L_order}. This can be understood as follows. Let us assume there are two points $\bx_1$ and $\bx_2$ on the inflated mesh whose closest surface point $\bx_0$ is the same, as illustrated by Fig. \ref{fig:reason-black}. Let us further assume that $\bx_2$ is initially farther from the body than $\bx_1$. After deformation, nothing prevents $\bx_1$ from ending up farther than $\bx_2$ and yielding a self-intersection. To prevent this, we introduce the ordering loss 
\begin{align}
  L_{order} &= \sum_{(\bx_1, \bx_2)\in O} max(0, SDF_{B}(\bx_2+\Delta x_{\theta}(\bx_2))-SDF_{B}(\bx_1+\Delta x_{\theta}(\bx_1))) \; ,\nonumber \\
  O&=\{(x_1, x_2)|\psi(x_2) =\psi(x_1) ~\text{and}~ SDF_{B}(x_1)>SDF_{B}(x_2)\} \; , \label{eq:order}
\end{align}
where $\psi(\cdot)$ denotes the closest garment vertex. Its minimization maintains the spatial relationship between points like $\bx_1$ and $\bx_2$ because it ensures that  points, close to the body before deformation are still close after deformation. 

\input{figs/L_order.tex}

\subsection{Implementation Details}

The SDF $f_{\Theta}(\bx,\bz)$ of Eq.~\ref{eq:sdf} is implemented by a 9-layer multilayer perceptron (MLP) with a skip connection from the input layer to the middle. We use Softplus as the activation function. The $\Theta$ weights and the latent code $z\in \mathbb{R}^{12}$ for each garment are optimized jointly during the training using a learning rate of 5e-4. 

We use the architecture of \cite{Bertiche21a} to implement the pose displacement network $\Delta x_{\theta}$ of Eq.~\ref{eq:model}. It comprises two MLP's with ReLU activation in-between. One encodes the pose $\theta$ to an embedding, and the other one predicts the blend matrices for the input point. The pose displacement is computed as the matrix product of the embedding and the blend matrices. The weight distribution $w(\cdot)$ of Eq.~\ref{eq:ours-weight} is implemented by an MLP with an extra Softmax layer at the end to normalize the output. $N=1000$ points are sampled to obtain the ground-truth $\bar{w}$ used to train $w$. We use the ADAM \cite{Kingma14a} optimizer with a learning rate of 1e-3 for the training of $w$ and $\Delta x_{\theta}$.

%% file: figs/sdf.tex
\begin{figure}[t]
    \centering
    \subfigure[]{\label{fig:inflate1}
    \includegraphics[height=3.3cm, width=4.8cm]{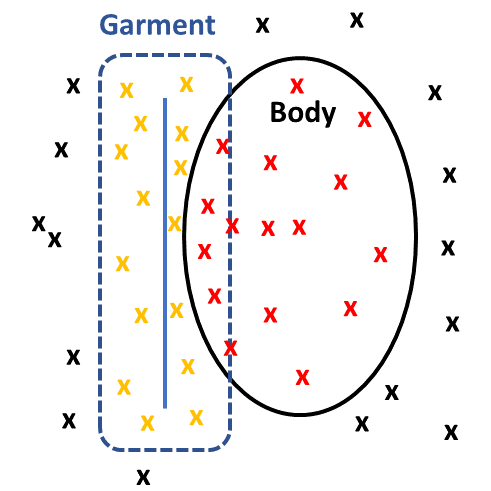}} ~~
    \subfigure[]{\label{fig:inflate2}
    \includegraphics[height=3.3cm, width=4.8cm]{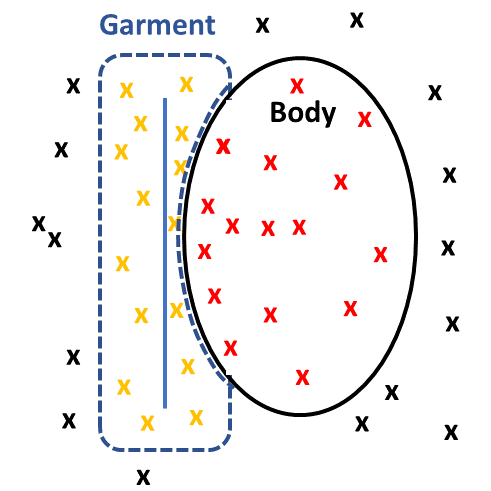}}
    \caption{The illustration of inflation processing for the garment surface (blue solid lines). (a) The inflation strategy of \cite{Guillard22a,Corona21} will cause interpenetration between the inflated mesh (blue dashed line) and the body mesh, while (b) our proposed interpenetration-aware inflation will not.}
    \label{fig:sdf}
\end{figure}

%% file: figs/L_order.tex
\begin{figure}[ht!]
    \centering
    \subfigure[]{\label{fig:reason-black}
    \includegraphics[width=0.47\linewidth]{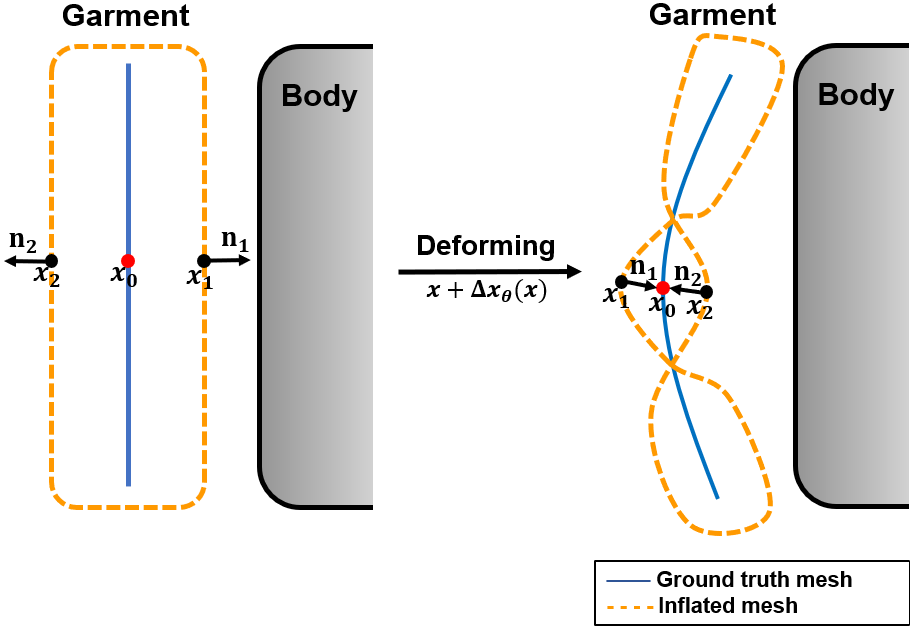}}~~~
    \subfigure[]{\label{fig:L_order}
    \includegraphics[width=0.34\linewidth]{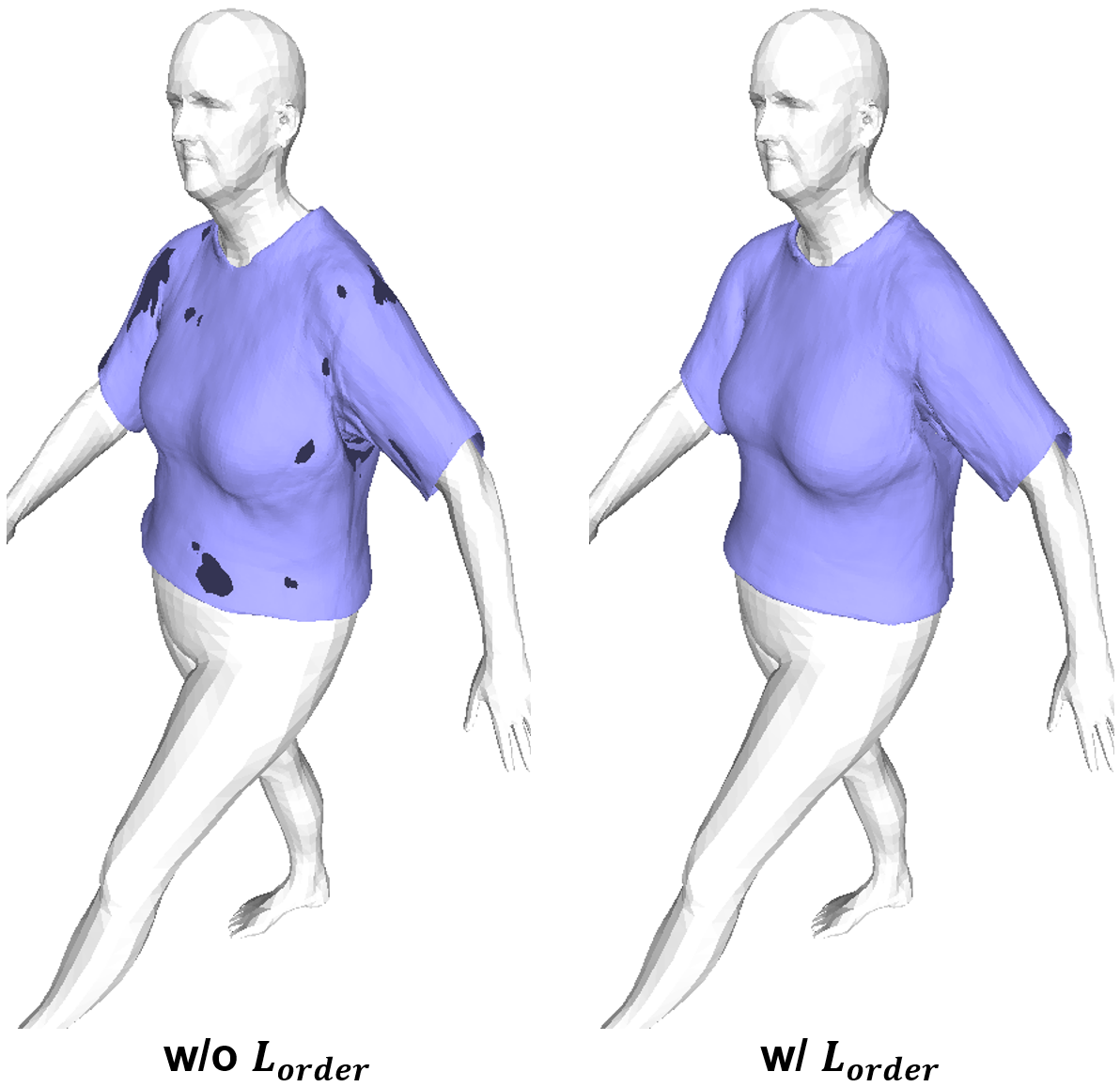}}
    \caption{(a) The illustration of how artifacts are produced by deformation. (b) Shirts deformed by the models trained w/ and w/o $L_{order}$. Without it, the inner face of the t-shirt can intersect the outer one. This is shown in dark blue.}
\end{figure}

%% file: tex/5_experiments.tex

\section{Experiments and Results}
\label{sec:expResults}

Our models can operate in several different ways. First, they can serve as generative models.  By varying the latent code of our SDF $f_{\Theta}$ and using Marching Cubes, we can generate triangulated surfaces for garments of different topologies that can then be draped over bodies of changing shapes and poses using the deformation model $M_G$ of Eq.~\ref{eq:model}, as shown in Fig. \ref{fig:teaser1}. Second, they can be used to recover both body {\it and} cloth shapes from images by minimizing
\begin{align}
  L(\beta,\theta,\bz) &= L_{\text{IoU}}(NeuR(M_G(\textbf{G},\beta,\theta), M_B(\beta,\theta)), \textbf{S}) + L_{prior}(\theta)\; , \label{eq:fit} \\
  \textbf{G} &= MC(f_{\Theta}(\bx,\bz)) \; , \nonumber
\end{align}
where $L_{\text{IoU}}$ is the IoU loss~\cite{Li21g} that measures the difference between segmentation masks, $\textbf{G}$ is the garment  surface reconstructed by Marching Cubes $MC(\cdot)$, $NeuR(\cdot)$ is a differentiable renderer, and $\textbf{S}$ a semantic segmentation obtained using off-the-shelf algorithms. $M_B(\cdot)$ and $M_G(\cdot)$ are the skinning functions for garment and body as defined in Eq. \ref{eq:smpl} and \ref{eq:model}. We also minimize the prior loss $L_{prior}$ of VPoser~\cite{Pavlakos19} to ensure plausibility of the pose.

In theory $MC(\cdot)$ is not differentiable, but its gradient at vertex $\bx$ can be approximated by $\frac{\partial \bx}{\partial \bz}=-\textbf{n}\frac{\partial f(\bx,\bz)}{\partial \bz}$, where $\textbf{n}=\nabla f(\bx)$ is the normal~\cite{Guillard22a}. In practice, this makes the minimization of Eq. \ref{eq:fit} practical using standard gradient-based tools and we again rely on ADAM. Pytorch3D \cite{Pytorch3D} serves as the differentiable renderer. In our experiments, we model shirts and trousers and use separate SDF and separate skinning models for each. Our pipeline can be used to jointly optimize the body mesh ($\beta$ and $\theta$) and the garment mesh ($\bz$), while previous work~\cite{Corona21} can only be used to optimize the garment. 

In this section, we demonstrate both uses of our model. To this end, we first introduce the dataset and metrics used for our experiments. We then evaluate our method and compare its performance with baselines for garment reconstruction and deformation. Finally, we demonstrate the ability of our method to model people and their clothes from synthetic and real images. 

\subsection{Dataset and Evaluation Metrics}

We train our models on data from CLOTH3D~\cite{bertiche20}. It contains over 7k sequences of different garments draped on animated 3D human SMPL models. Each garment has a different template and a single motion sequence that is up to 300 frames long. We randomly select 100 shirts and 100 trousers, and transform them to a body with neutral-shape and T-pose by using displacement of the closest SMPL body vertex, which yields meshes in the canonical space. For each garment sequence, we use the first 90\% frames as the training data and the rest as the test data (denoted as TEST EASY). We also randomly select 30 unseen sequences (denoted as TEST HARD) to test the generalization ability of our model. Chamfer Distance (CD), Euclidean Distance (ED), Normal Consistency (NC) and Interpenetration Ratio (IR) are reported as the evaluation metrics. NC is implemented as in~\cite{Guillard22a}. IR is computed as the area ratio of garment faces inside the body to the overall garment faces.

\subsection{Garment Reconstruction}
\label{sec:expRecon}

The insets of Fig.~\ref{fig:collision} contrast our reconstruction results against those of SMPLicit~\cite{Corona21}. The latter yields large interpenetrations while the former does not. Fig. \ref{fig:L_grad} showcases the role of the $L_{grad}$ term of Eq.~\ref{eq:sdf-grad} in producing smooth surfaces.

In Table \ref{tab:recon}, we report quantitative results for both the shirt and the trousers. We outperform SMPLicit (the first row - w/o \textit{proc.}, w/o $L_{grad}$) in all three metrics. The margin in IR is over 18\%, which showcases the ability of the interpenetration-aware processing of Section~\ref{sec:recon}.

\input{figs/collision.tex}
\input{tables/reconstruction.tex}
\input{tables/deform.tex}

\subsection{Garment Deformation}
\label{sec:expDeform}
\input{tables/deform-unseen.tex}
\input{tables/deform-sdf.tex}
In this section, we compare our deformation results against those of SMPLicit~\cite{Corona21} and DeePSD~\cite{Bertiche21}. The input to DeePSD is the point cloud formed by the vertices of ground-truth mesh so that, like our algorithm, it can deform garments of arbitrary topology by estimating the deformation for each point separately. To skin the garment, it learns functions to predict the blending weight and pose displacement. It also includes a self-consistency module to handle body-garment interpenetration. Hence, for a fair comparison, we retrain DeePSD using the same training data as before.

To test the deformation behavior of our model, we use the SMPL parameters $\beta$ and $\theta$ provided by the test data as the input of our skinning model. As to the garment mesh to be deformed, we either use the ground-truth unposed mesh from the data, which is an open surface, or the corresponding watertight mesh reconstructed by our SDF model. 

In Fig.~\ref{fig:L_order}, we presented a qualitative result that shows the importance of the ordering term of $L_{order}$ in Eq.~\ref{eq:order}. We report quantitative results with the  ground-truth mesh in Table~\ref{tab:deform} on TEST EASY. Our model performs substantially better than both baselines with the lowest ED and IR and the highest NC. For example, comparing to SMPLicit, the ED and IR of our model drop by more than 15mm and 10\% for the deformation of shirt. In Table~\ref{tab:deform-unseen}, we report similar results on TEST HARD, which is more challenging since it resembles less the training set, and we can draw the same conclusions. Since the learning of blending weights in DeePSD does not exploit the prior of the body model as us (Eq. \ref{eq:ours-weight}), it suffers a huge performance deterioration in this case where its ED even goes up to 95.6mm and 37.8mm for the shirt and trousers respectively. Table \ref{tab:deform-sdf} reports the results with SDF reconstructed mesh. Again, our method performs consistently better than SMPLicit in all metrics (row 2 vs row 4). It is also noteworthy that our interpenetration-aware pre-processing can help reduce deformation error and interpenetration ratio as indicated by the results of row 3 and 4. This demonstrates that learning a physically accurate model of garment interpenetrations results in more accurate clothing deformations.

In the qualitative results of Fig. \ref{fig:deform}, we can observe that SMPLicit cannot generate realistic dynamics and its results tend to be over-smoothed due to its simple skinning strategy. DeePSD can produce results that are better but too noisy. Besides, neither of them is able to address the body-garment interpenetration. Fig. \ref{fig:heatmap} visualizes the level of interpenetrations happening different body region. We can notice that interpenetrations occur on almost everywhere in the body for SMPLicit. DeePSD shows less but still not as good as ours.
\input{figs/deform.tex}
\input{figs/heatmap.tex}

\input{tables/synthetic.tex}
\subsection{From Images to Clothed People}
\label{sec:expFit}

Our model can be used to recover the body and garment shapes of clothed people from images by minimizing $L$ of Eq. \ref{eq:fit} with respect to $\beta$, $\theta$, and $\bz$. To demonstrate this, we use both synthetic and real images and compare our results to those of SMPLicit. Our optimizer directly uses the posed garment to compute the loss terms. In contrast, SMPLicit performs the optimization on the unposed garment. It first samples 3D points $\textbf{p}$ in the canonical space. i.e. on the unposed body, and uses the weights of the closest body vertices to project these points into posed space and into 2D image space to determine if semantic label, 1 if inside the garment, 0 otherwise.  The loss 
\begin{equation}\label{eq:smplicit}
    L(\bz_G) = \begin{cases}
              |C(\textbf{p},\bz_G)-\textbf{d}_{max}|, & \text{if}~ s_\textbf{p}=0\\ 
              \text{min}_i|C(\textbf{p}^i,\bz_G)|, & \text{if}~ s_\textbf{p}=1
             \end{cases},
\end{equation}
is then minimized with respect to the latent code $\bz_G$, where $\textbf{d}_{max}$ is the maximum cut-off distance and $\text{min}_i$ is used to consider only the point closest to the current garment surface estimate. This fairly complex processing chain tends to introduce inaccuracies.

\paragraph{\bf Synthetic Images.} 
\input{figs/synthetic.tex}
We use the body and garment meshes from CLOTH3D as the synthetic data. Since the ground-truth SMPL parameters are available, we only optimize the latent code $\bz$ for the garment  
and drop  the pose prior term $L_{prior}$ from Eq. \ref{eq:fit}. Image segmentation such as the one of Fig.~\ref{fig:synthetic} are obtained by using Pytorch3D to render meshes under specific camera configurations. Given the ground-truth $\beta$, $\theta$ and segmentation, we initialize $\bz$ as the mean of learned codes and then minimize the loss. Fig \ref{fig:synthetic} shows qualitative results in one specific case. The quantitative results reported in Table~\ref{tab:synthetic} confirm the greater accuracy and lesser propensity to produce interpenetrations of our approach.

\paragraph{\bf Real Images.} 
\input{figs/real.tex}
In real-world scenarios such as those depicted by  Fig.~\ref{fig:in-the-wild}, there are no ground-truth annotations but we can get the required information from single images from off-the-shelf algorithms. As in SMPLicit, we use~\cite{rong20} to estimate the SMPL parameters $\hat{\beta}$ and $\hat{\theta}$  and the algorithm of~\cite{yang20} to produce a segmentation. In SMPLicit, $\hat{\beta}$ and $\hat{\theta}$ are fixed and only the garment model is updated. In contrast, in our approach, $\hat{\beta}$, $\hat{\theta}$, and the latent vector $\bz$ are all optimized. As can be seen in Fig.~\ref{fig:in-the-wild}, this means that inaccuracies in the $\hat{\beta}$ and $\hat{\theta}$ initial values can be corrected, resulting in an overall better fit of both body and garments.

%% file: figs/collision.tex

\begin{figure}[t]
    \centering
    \subfigure[]{\label{fig:collision}
    \includegraphics[width=.34\textwidth]{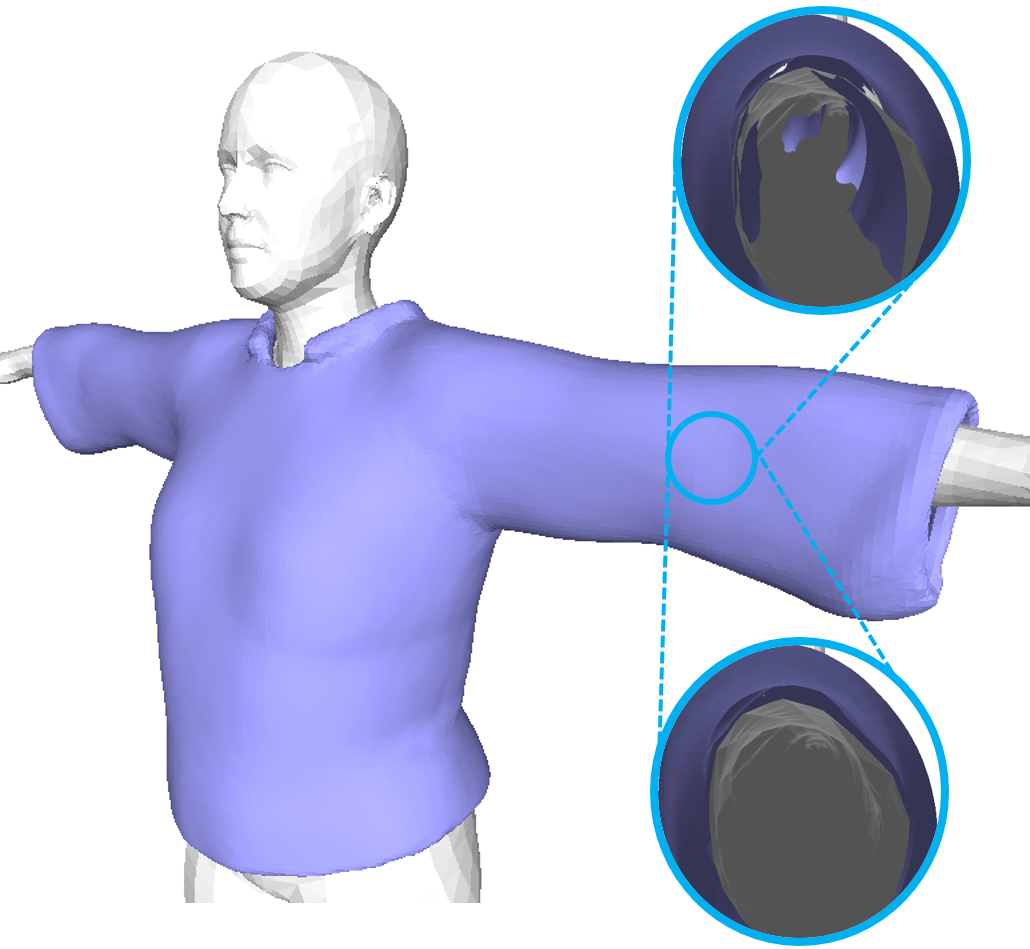}}~~
    \subfigure[]{\label{fig:L_grad}
    \includegraphics[width=.46\textwidth]{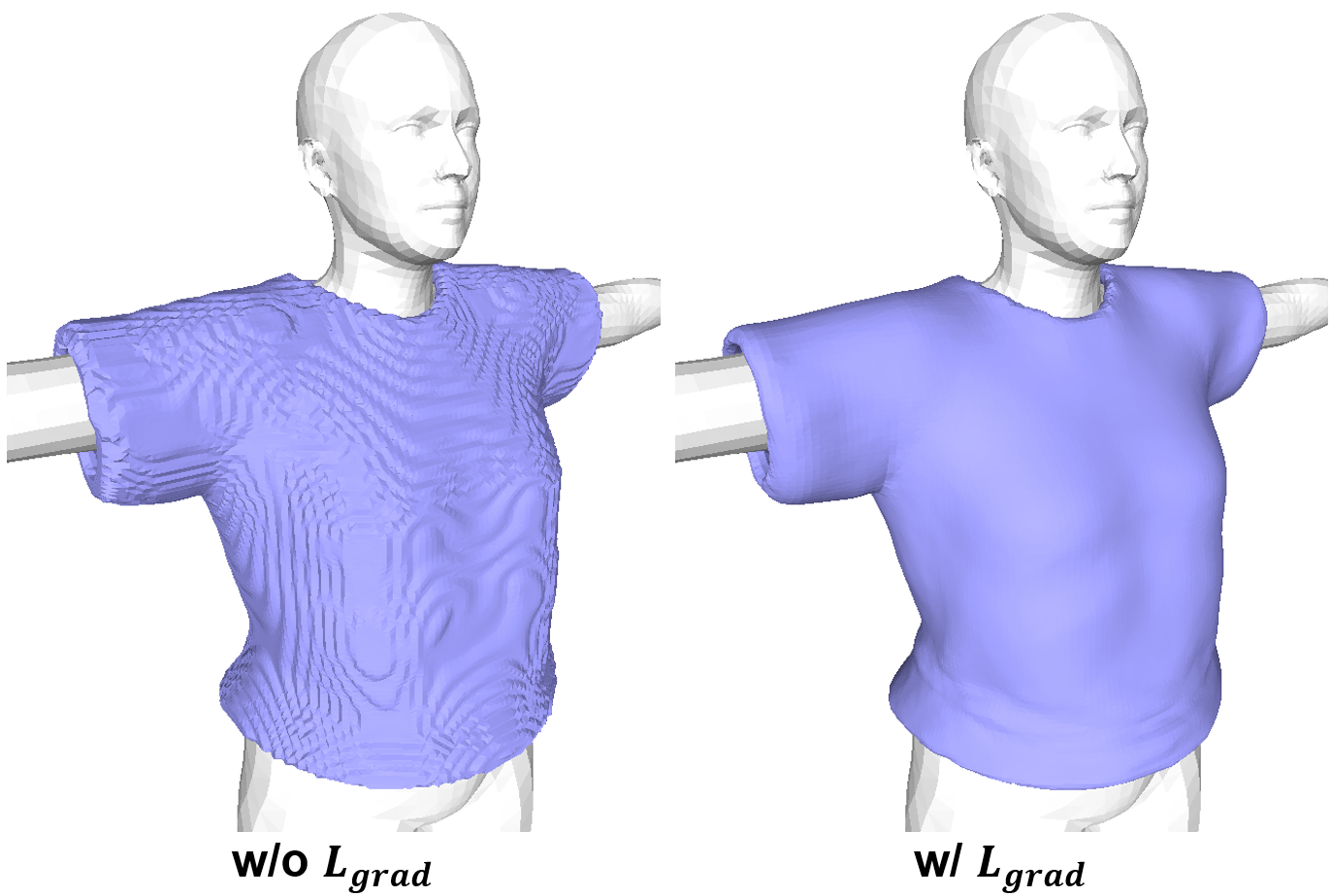}}
    \caption{{\bf Reconstruction results.} (a) The inside of the same garment reconstructed by SMPLicit (upper inset) and our method (bottom inset). The first features interpenetrations whereas the second does not. (b) Reconstructed garment by a model trained without and with $L_{grad}$. The latter is smoother and preserves details better.}
  \label{fig:reconstruction}
\end{figure}

%% file: tables/reconstruction.tex

\begin{table}[h!]
  \begin{center}
  \scalebox{.7}{
    \begin{tabular}{c | c | c | c}
    \toprule
    Shirt & CD ($\times 10^{-4}$) & NC (\%)  &IR (\%) \\
    \midrule
    Ours w/o \textit{proc.}, w/o $L_{grad}$& 1.88          & 92.1          & 18.1 \\
    Ours w/o $L_{grad}$  & 1.58          & 90.3          & \textbf{0.0} \\
    Ours   & \textbf{1.48} & \textbf{92.3} & \textbf{0.0} \\
    \bottomrule
    \end{tabular}
  }
  \scalebox{.7}{
    \begin{tabular}{c | c | c | c}
    \toprule
    Trousers & CD ($\times 10^{-4}$) & NC (\%)  &IR (\%) \\
    \midrule
    Ours w/o \textit{proc.}, w/o $L_{grad}$               & 1.65          & 92.0          & 18.6 \\
    Ours w/o $L_{grad}$  & \textbf{1.22} & 91.8          & \textbf{0.0} \\
    Ours   & 1.34          & \textbf{92.3} & \textbf{0.0} \\
    \bottomrule
    \end{tabular}
    }
  \end{center}
  \caption{Comparative reconstruction results. \textit{proc.} indicates our proposed interpenetration-aware pre-processing.}
    \label{tab:recon}
\end{table}

%% file: tables/deform.tex
\begin{table}[h!]
    \begin{center}
      \scalebox{.85}{
        \begin{tabular}{c | c | c | c}
        \toprule
        Shirt & ED (mm) & NC (\%)  &IR (\%) \\
        \midrule
        DeePSD       & 26.1          & 82.3          & 5.8 \\
         SMPLicit   & 35.9          & 84.0          & 13.3 \\
         Ours       & \textbf{19.0} & \textbf{85.3} & \textbf{1.6} \\
        \bottomrule
        \end{tabular}
      }
      \scalebox{.85}{
        \begin{tabular}{c | c | c | c}
        \toprule
        Trousers & ED (mm) & NC (\%)  &IR (\%) \\
        \midrule
        DeePSD   & 17.5          & 85.4          & 1.5 \\
         SMPLicit   & 27.0          & 85.6          & 6.3 \\
         Ours       & \textbf{14.8} & \textbf{86.7} & \textbf{0.2} \\
        \bottomrule
        \end{tabular}
        }
      \end{center}
      \caption{Deforming unposed ground truth garments with DeePSD, SMPLicit and our method on TEST EASY.}
      \label{tab:deform}
\end{table}

%% file: tables/deform-unseen.tex
\begin{table}[h!]
    \begin{center}
      \scalebox{.85}{
        \begin{tabular}{c | c | c | c}
        \toprule
        Shirt & ED (mm) & NC (\%)  &IR (\%) \\
        \midrule
        DeePSD   & 95.6          & 72.6          & 46.4 \\
        SMPLicit   & 35.4          & 83.9          & 12.9 \\
        Ours       & \textbf{26.5} & \textbf{85.1} & \textbf{3.0} \\
        \bottomrule
        \end{tabular}
      }
      \scalebox{.85}{
        \begin{tabular}{c | c | c | c}
        \toprule
        Trousers & ED (mm) & NC (\%)  &IR (\%) \\
        \midrule
        DeePSD   & 37.8         & 79.5          & 27.8 \\
        SMPLicit  & 31.9          & 84.9          & 8.8 \\
        Ours      & \textbf{24.8} & \textbf{85.8} & \textbf{0.7} \\
        \bottomrule
        \end{tabular}
        }
      \end{center}
      \caption{Deforming unposed ground truth garments with DeePSD, SMPLicit and our method on TEST HARD.}
      \label{tab:deform-unseen}
\end{table}

%% file: tables/deform-sdf.tex
\begin{table}[h!]
    \begin{center}
      \scalebox{.8}{
        \begin{tabular}{c | c | c | c}
        \toprule
        Shirt & CD ($\times 10^{-4}$) & NC (\%)  &IR (\%) \\
        \midrule
         SMPLicit   & 7.91          & 83.7          & 16.2 \\
         \midrule
         Ours - w/o \textit{proc.}       & 3.86 & 84.4 & 1.6 \\
         Ours - w/ \textit{proc.}      & \textbf{3.78} & \textbf{84.7} & \textbf{1.5} \\
        \bottomrule
        \end{tabular}
      }
      \scalebox{.8}{
        \begin{tabular}{c | c | c | c}
        \toprule
        Trousers & CD ($\times 10^{-4}$) & NC (\%)  &IR (\%) \\
        \midrule
         SMPLicit   & 3.66          & 84.1          & 6.6 \\
         \midrule
         Ours - w/o \textit{proc.}        & 2.71          & 85.3          & \textbf{0.2} \\
         Ours - w/ \textit{proc.}      & \textbf{2.67} & \textbf{85.4} & \textbf{0.2} \\
        \bottomrule
        \end{tabular}
        }
      \end{center}
      \caption{Deforming SDF reconstructed garments with SMPLicit and our method on TEST EASY. w/o and w/ \textit{proc.} means the mesh is reconstructed without and with interpenetration-aware processing respectively.}
      \label{tab:deform-sdf}
\end{table}

%% file: figs/deform.tex
\begin{figure}[t]
    \centering
    \includegraphics[width=.9\linewidth]{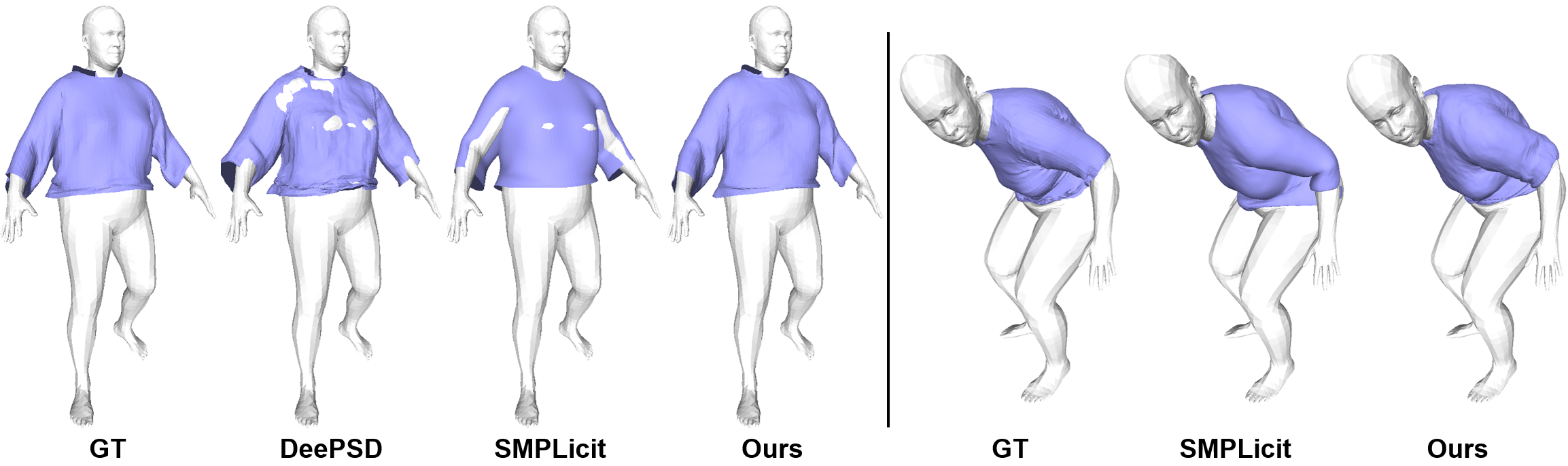}
    \caption{The skinning results for the ground-truth shirt (left) and the SDF reconstructed shirt (right). Since the input of DeePSD should be the point cloud of the mesh template, we only evaluate it with the unposed ground-truth mesh. Compared to DeePSD and SMPLicit, our method can produce more realistic details and have less body-garment interpenetration.}
    \label{fig:deform}
\end{figure}

%% file: figs/heatmap.tex
\begin{figure}[t]
    \centering
    \subfigure[DeePSD]{\label{fig:heatmap-DeePSD}
    \includegraphics[width=0.26\linewidth]{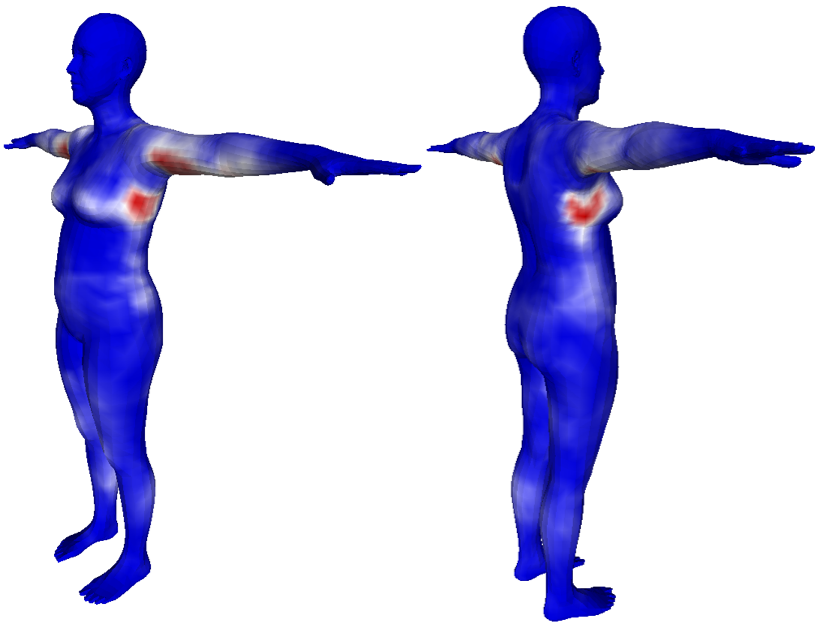}}~~~
    \subfigure[SMPLicit]{\label{fig:heatmap-SMPLicit}
    \includegraphics[width=0.26\linewidth]{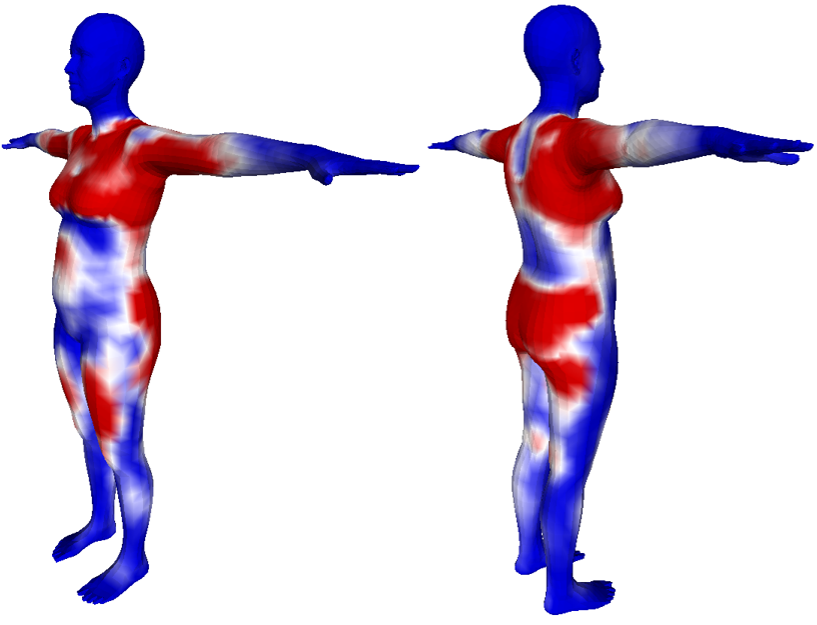}}~~~
    \subfigure[Ours]{\label{fig:heatmap-ours}
    \includegraphics[width=0.26\linewidth]{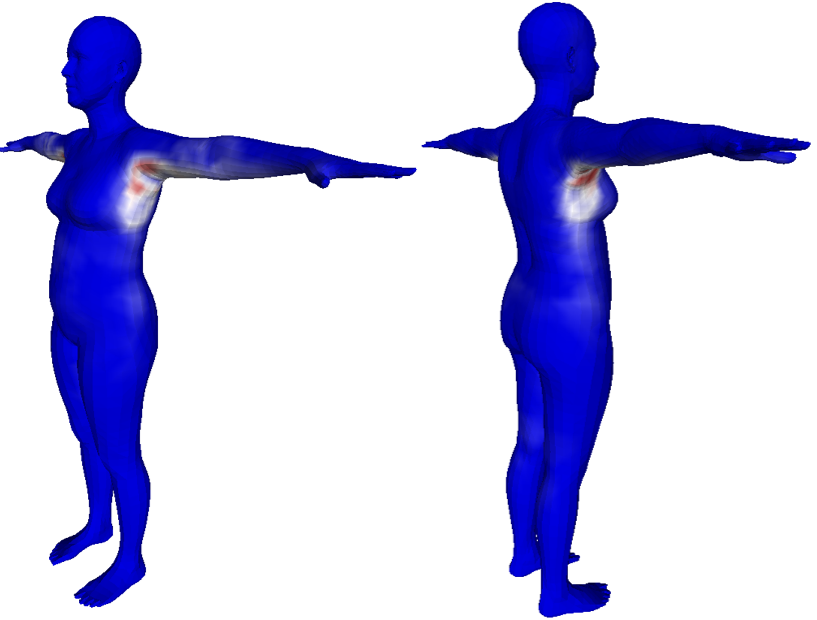}}
  \caption{The visualization of the body region having interpenetrations (marked in red).}
  \label{fig:heatmap}
\end{figure}

%% file: tables/synthetic.tex
\begin{table}[h!]
    \begin{center}
    \scalebox{.8}{
      \begin{tabular}{c | c | c | c}
      \toprule
      Shirt & CD ($\times 10^{-4}$) & NC (\%)  &IR (\%) \\
      \midrule
       SMPLicit-raw               & 17.77          & 82.1          & 41.5 \\
       SMPLicit  & 18.73          & 82.8          & 37.3 \\
      \midrule
       Ours   & \textbf{4.69} & \textbf{87.3} & \textbf{3.9} \\
      \bottomrule
      \end{tabular}
    }
    \scalebox{.8}{
      \begin{tabular}{c | c | c | c}
      \toprule
      Trousers & CD ($\times 10^{-4}$) & NC (\%)  &IR (\%) \\
      \midrule
       SMPLicit-raw               & 4.22          & 81.2          & 35.7 \\
       SMPLicit  & 4.50          & 82.2          & 29.2 \\
       \midrule
       Ours   & \textbf{2.23} & \textbf{89.2} & \textbf{0.7} \\
      \bottomrule
      \end{tabular}
      }
    \end{center}
    \caption{The evaluation results of SMPLicit-raw (w/o smoothing), SMPLicit (w/ smoothing) and our method for garment fitting on the synthetic data.}
      \label{tab:synthetic}
\end{table}

%% file: figs/synthetic.tex
\begin{figure}[ht]
    \centering
    \includegraphics[width=.8\linewidth]{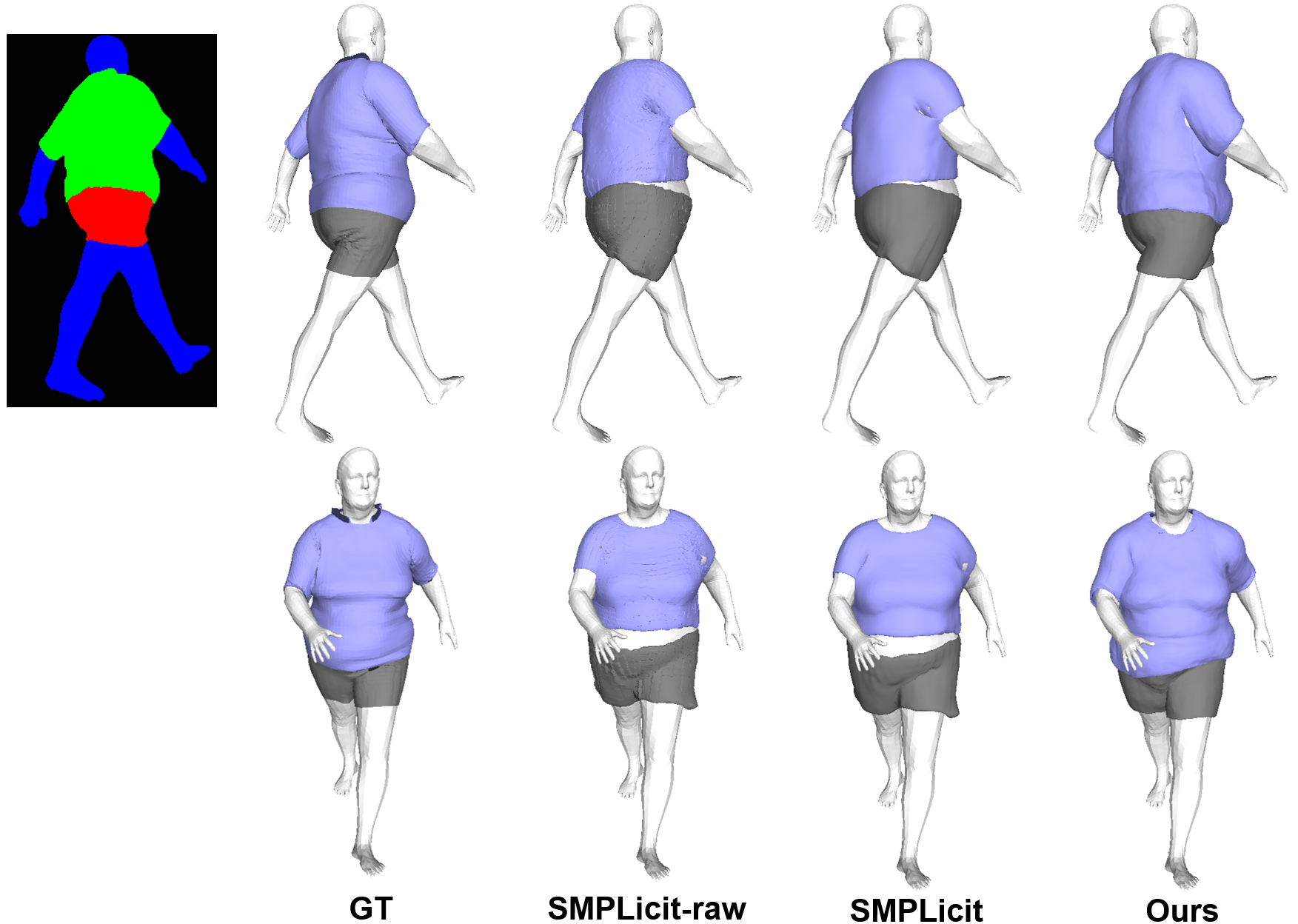}
    \caption{{\bf Fitting results on a synthetic image.} Left to right: the ground-truth segmentation and garment meshes, SMPLicit w/o smoothing (SMPLicit-raw), SMPLicit w/ smoothing (SMPLicit) and ours. Note that SMPLicit requires post-processing to remove artifacts, while our method does not.}
    \label{fig:synthetic}
\end{figure}

%% file: figs/real.tex
\begin{figure}[ht!]
    \centering
    \includegraphics[width=1.\linewidth]{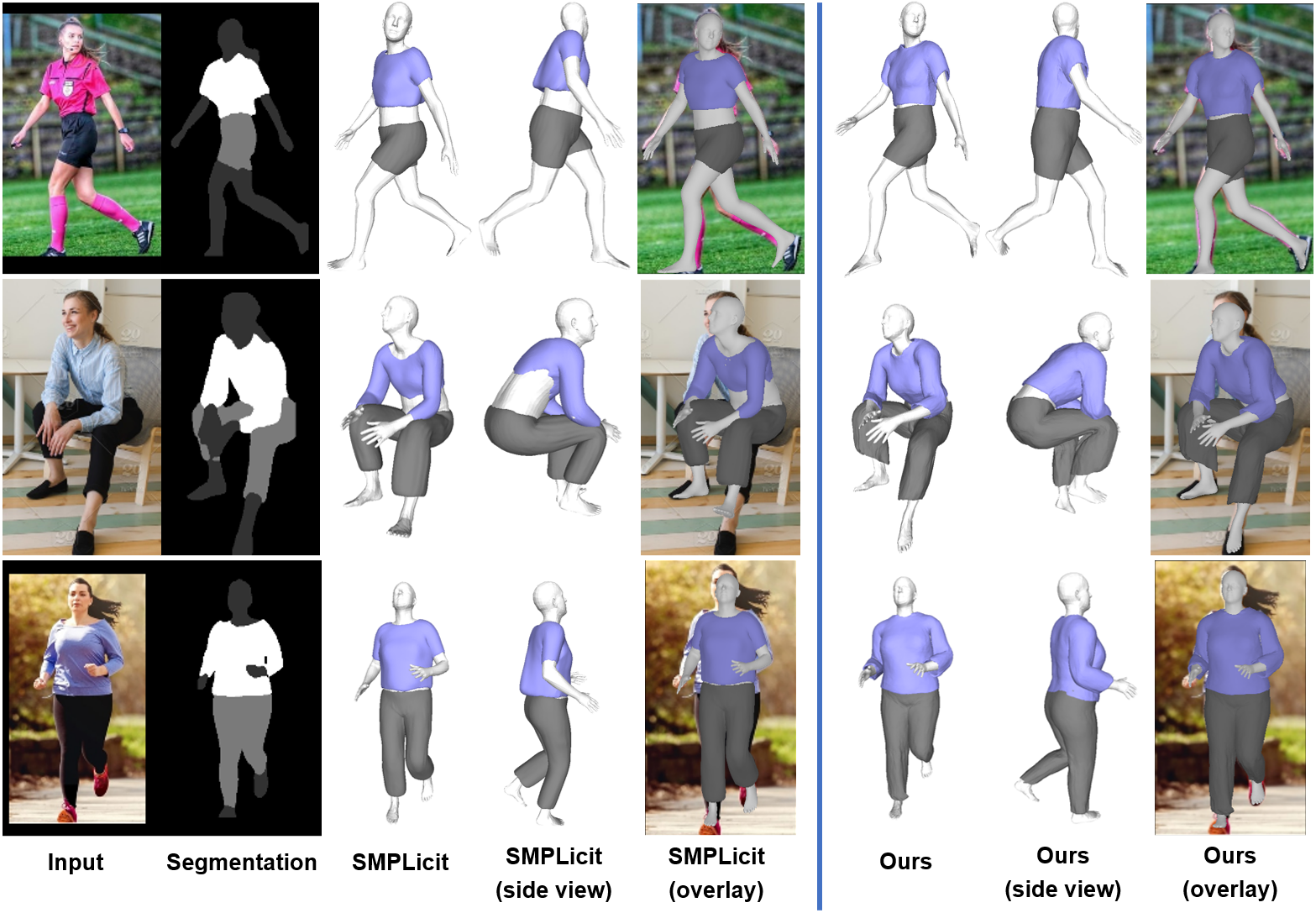}
    \caption{\textbf{Fitting results on images in-the-wild.} Left to right: the input images and their segmentation, SMPLicit and ours. Note that SMPLicit recovers garments based on body meshes estimated from \cite{rong20}, while we can optimize the body and garment parameters jointly for more accurate results.}
    \label{fig:in-the-wild}
\end{figure}

%% file: tex/6_conclusion.tex

\section{Conclusion}

We have presented a fully differentiable approach to draping a garment on a body so that both body and garment parameters can be jointly optimized. At its heart is a skinning model that learns to prevent self-penetration. We have demonstrated its effectiveness both for animation purposes and to recover body and cloth shapes from real images. In future work, we will incorporate additional physics-based constraints to increase realism and to reduce the required amount of training data. 

%% file: tex/7_supp.tex
\section{Dataset Pre-Processing}

The garment mesh template in CLOTH3D~\cite{bertiche20} is designed for T-pose body with shape $\beta \ne 0$. To transform it back to the body with neutral shape ($\beta=0$), for each garment vertex $\bv_g$, we first find its closest body vertex $\bv_b$ and get the corresponding shape displacement $B_s(\bv_b,\beta)$ from SMPL~\cite{Loper14}, and then $\bv_g$ is replaced by the new position $\bv_g-B_s(\bv_b,\beta)$, which gives us the garment on the neutral body. We also apply a Laplacian operator with $\lambda=0.2$ afterwards to smooth the transformed garment.

\section{Weight coefficients}
Following \cite{Saito12}, we use $\lambda_{grad}=0.1$ and $\lambda_{reg}=0.001$ in Eq. 2. In Eq. 11, we use $\lambda_{deform}=1$, $\lambda_{interp}=0.1$ and $\lambda_{order}=0.01$. When the values of $\lambda_{interp}$ and $\lambda_{order}$ are too large, the deformation accuracy decreases because of the trade-off between loss terms. Moreover, using a larger value for $\lambda_{interp}$ does not prevent interpenetration around the armpits as shown in the main paper Fig. 7(c) since arms can get very close to the body for some poses and self-collision can even happen on the GT body mesh as illustrated by Fig. \ref{fig:limit1}.

\section{Additional Evaluation}
\subsection{Examples of Reconstruction}

In Fig.~\ref{fig:sdf-shirt} and \ref{fig:sdf-trousers}, we show some samples of shirts and trousers reconstructed by our SDF model.
\input{figs/sdf-shirt.tex}
\input{figs/sdf-trousers.tex}

\subsection{Ablation Study of $L_{order}$}

In Table. \ref{tab:deform-order}, we compare the deformation results for our models trained with and without $L_{order}$. Using $L_{order}$ can decrease the deformation accuracy but the margin is small. Considering the efficacy of $L_{order}$ in resolving garment self-intersections as illustrated in Fig. 4(b) of the main paper, it is still necessary for us to have it for the better visual quality.
\input{tables/deform-order.tex}
\input{figs/deform_t.tex}

\subsection{Qualitative Results of Deformation}
In Fig.~\ref{fig:deform-t}, we show the deformation results of DeePSD~\cite{Bertiche21}, SMPLicit~\cite{Corona21} and ours for the trousers. Similar to the results on shirts, our method outperforms the baselines and shows more realistic details and less body-garment interpenetrations. 

Fig.~\ref{fig:supp-shape} shows the interpolation of the body shape with the same shirt. Notice that our method can produce reasonable deformation results for bodies with fat/slim and short/tall figures. Fig.~\ref{fig:supp-gar} shows the draping results of different shirts (randomly generated by our SDF model) on body in different shapes. The reconstructed shirts have different styles and lengths, but our model is still able to produce natural dynamics for them.
\input{figs/shape.tex}
\input{figs/garment.tex}

\subsection{The Accuracy of Body Reconstruction}
We conducted an experiment to measure the accuracy of body reconstruction. We use the data from CLOTH3D \cite{bertiche20} since it has the ground-truth body and garment mesh. We run \cite{rong20} to estimate the SMPL parameters $\hat{\beta}$ and $\hat{\theta}$ for the input image, and use them as the initial values for $\beta$ and $\theta$ to perform the optimization of Eq. 15. We use Mean Vertex Error (MVE) to measure the error between the ground-truth body mesh and the reconstruction. The MVE of the initial body estimated by \cite{rong20} is 127.2mm, while the MVE of our recovered mesh is 43.2mm. This demonstrates that our optimization can correct inaccuracies in the initial body mesh.

\section{Limitations}
\input{figs/limitation.tex}
Body-garment interpenetration can happen around armpits due to the small distance between arms and the body under specific poses (Fig. \ref{fig:limit1}). The performance of our model is a function of the mesh data used in training, so it is hard for us to recover the garment type not included in the training set (e.g., dress as Fig. \ref{fig:limit2}). Besides, our fitting strategy requires segmentation masks estimated by the segmentation algorithm (e.g., \cite{yang20}), which can fail under challenging scenarios and is not able to resolve depth ambiguity for pose as shown in Fig. \ref{fig:limit2} and \ref{fig:limit3}. In the future we plan to address interpenetration at armpits by operating in coordinates relative to the underlying SMPL template and depth ambiguity by considering  multi-view or depth observations.

%% file: figs/sdf-shirt.tex
\begin{figure}[ht!]
    \centering
    \includegraphics[width=1.\linewidth]{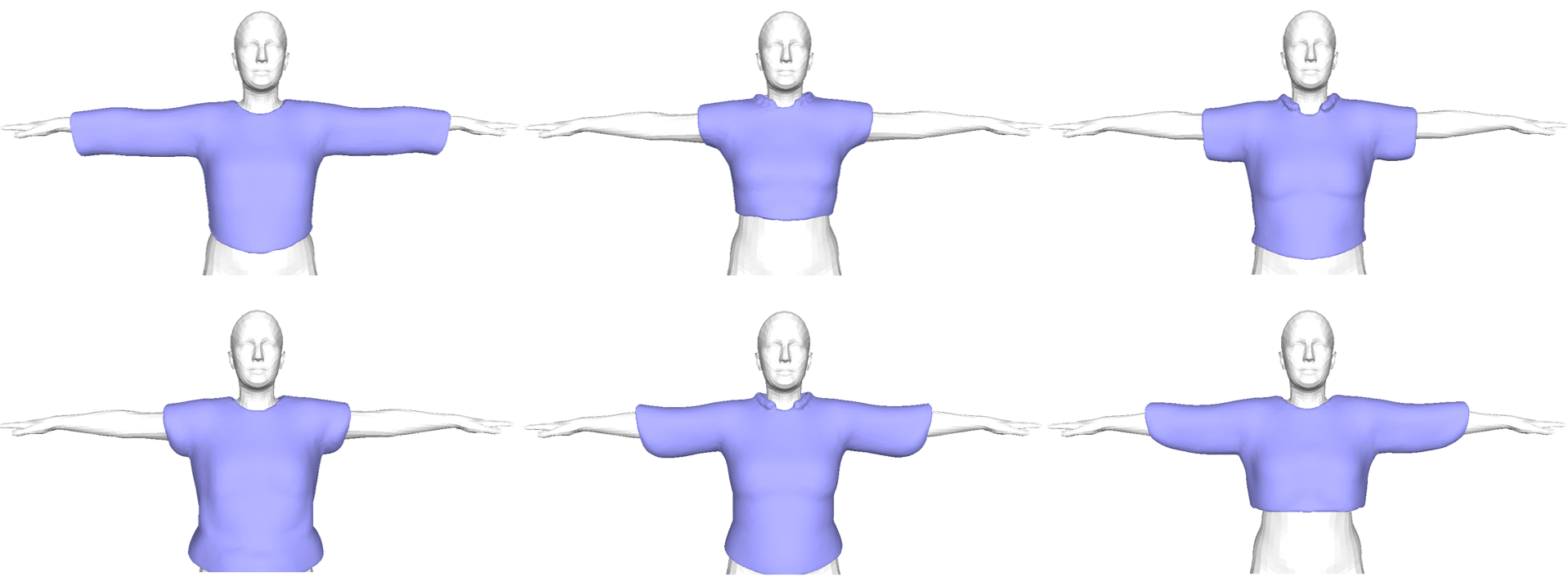}
    \caption{The samples of shirts reconstructed by our SDF model.}
    \label{fig:sdf-shirt}
\end{figure}

%% file: figs/sdf-trousers.tex
\begin{figure}[ht!]
    \centering
    \includegraphics[width=.9\linewidth]{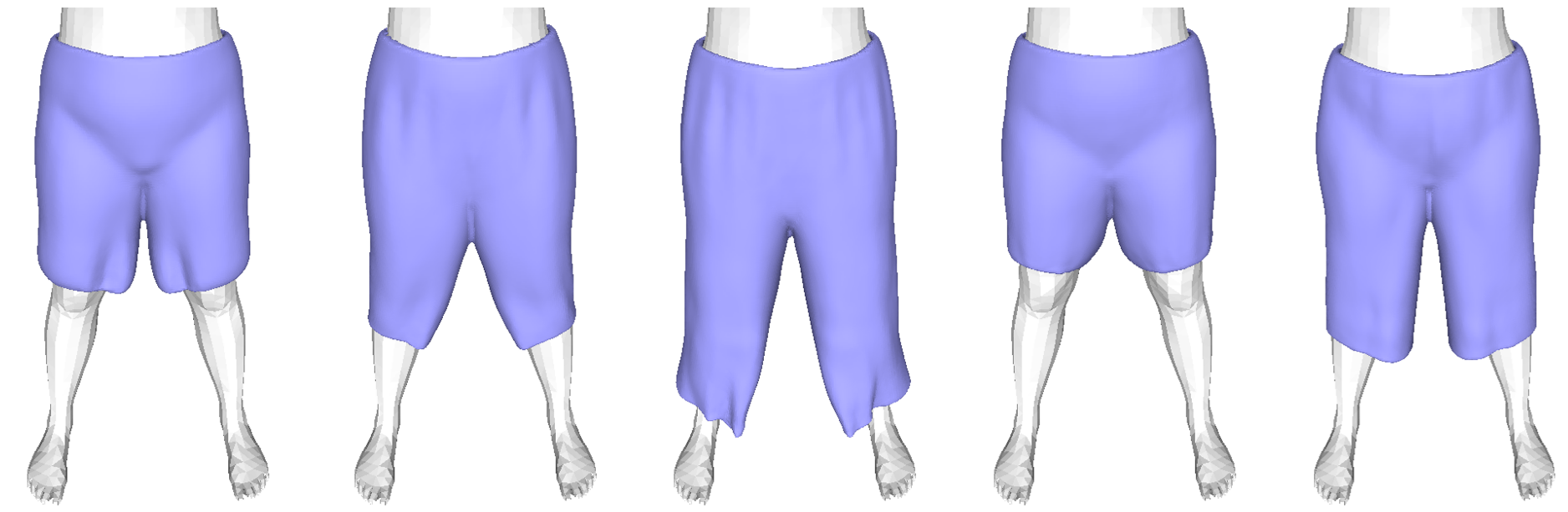}
    \caption{The samples of trousers reconstructed by our SDF model.}
    \label{fig:sdf-trousers}
\end{figure}

%% file: tables/deform-order.tex
\begin{table}[h!]
    \begin{center}
      \scalebox{1.}{
        \begin{tabular}{c | c | c | c}
        \toprule
        Shirt & CD ($\times 10^{-4}$) & NC (\%)  &IR (\%) \\
        \midrule
        w/o $L_{order}$    & \textbf{2.62} & \textbf{87.3} & \textbf{1.0} \\
        w/ $L_{order}$     & 3.78 & 84.7 & 1.5 \\
        \bottomrule
        \end{tabular}
      }
      \end{center}
      \caption{The evaluation results of our deformation models trained w/ and w/o $L_{order}$ on test sequences.}
      \label{tab:deform-order}
\end{table}

%% file: figs/deform_t.tex
\begin{figure}[ht!]
    \centering
    \includegraphics[width=1.\linewidth]{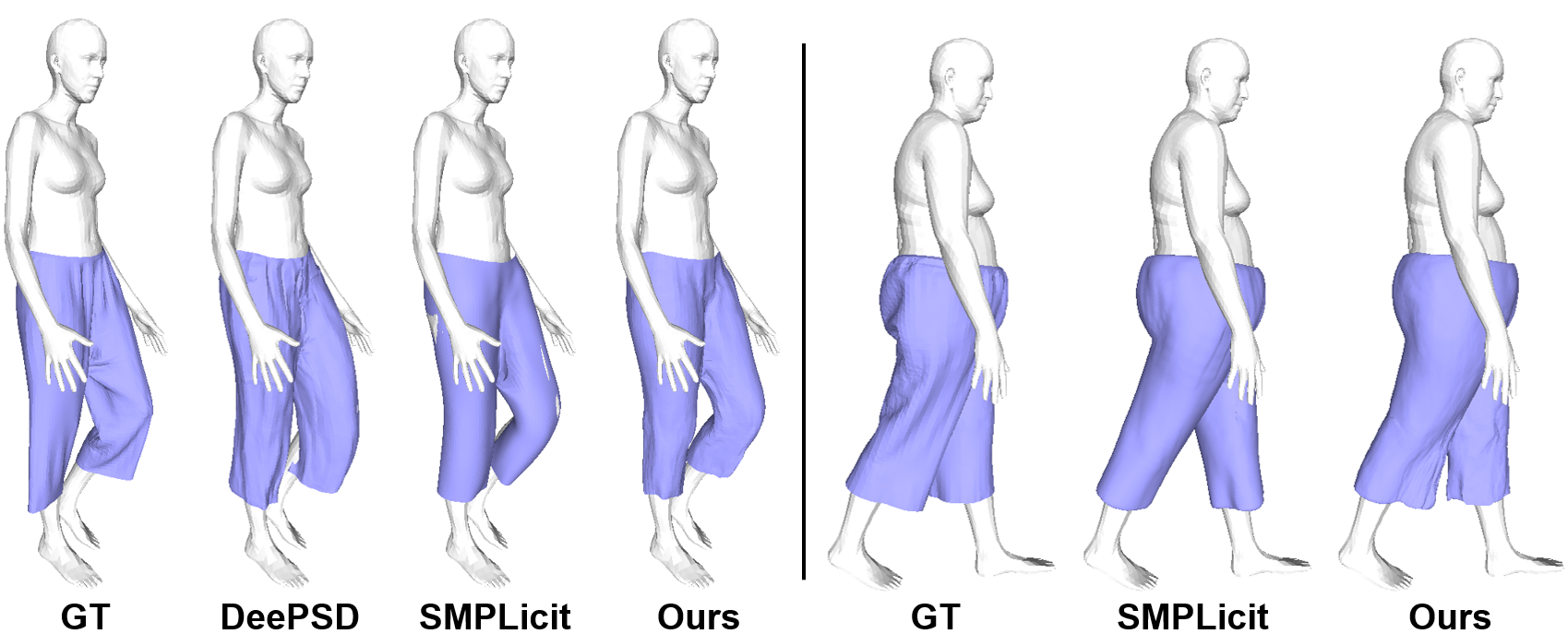}
    \caption{The skinning results for the ground-truth trousers (left) and the SDF reconstructed trousers (right). Since the input of DeePSD should be the point cloud of the mesh template, we only evaluate it with the unposed ground-truth mesh. Compared to DeePSD and SMPLicit, our method can produce more realistic details and have less body-garment interpenetrations.}
    \label{fig:deform-t}
\end{figure}

%% file: figs/shape.tex
\begin{figure}[ht!]
    \centering
    \includegraphics[width=1.\linewidth]{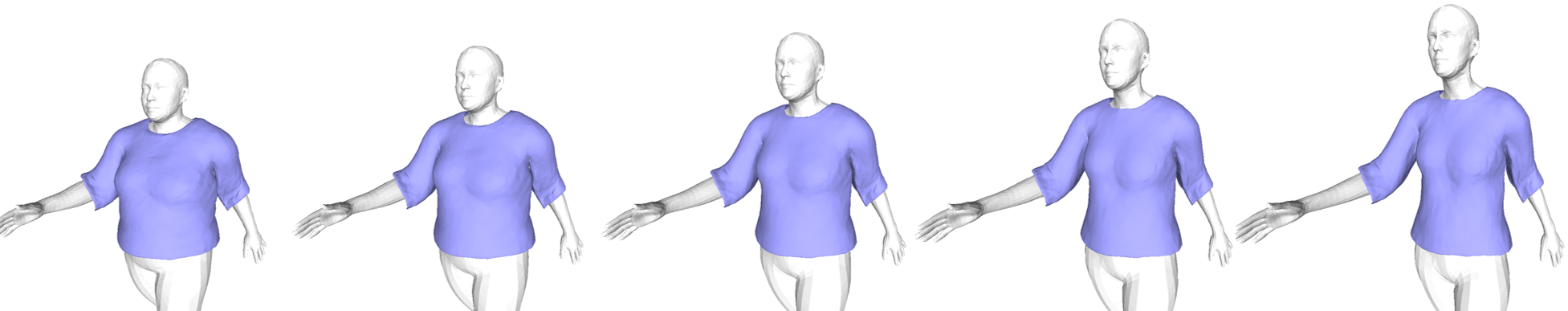}
    \caption{The draping results of the same shirt on body with different shapes.}
    \label{fig:supp-shape}
\end{figure}

%% file: figs/garment.tex
\begin{figure}[ht!]
    \centering
    \includegraphics[width=1.\linewidth]{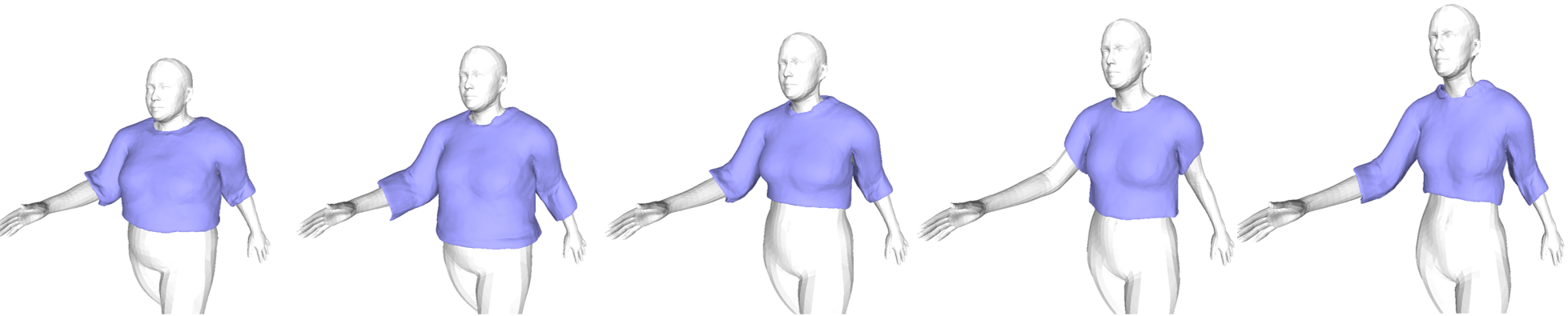}
    \caption{The draping results of different shirts, randomly generated by our SDF model, on body with different shapes.}
    \label{fig:supp-gar}
\end{figure}

%% file: figs/limitation.tex
\begin{figure}[ht]
    \centering
    \subfigure[]{\label{fig:limit1}
    \includegraphics[height=4cm]{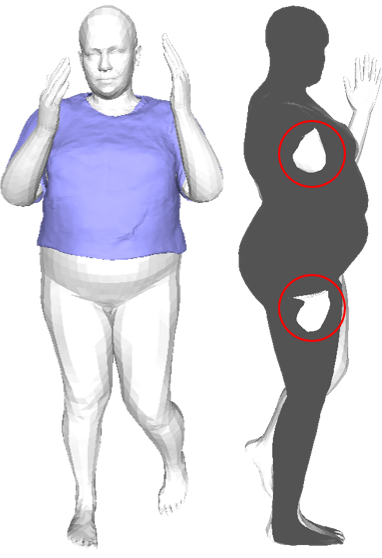}}~~
    \subfigure[]{\label{fig:limit2}
    \includegraphics[height=4cm]{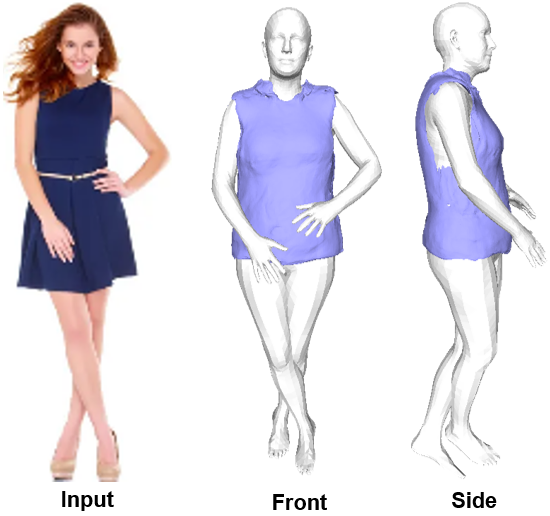}}~~
    \subfigure[]{\label{fig:limit3}
    \includegraphics[height=4cm]{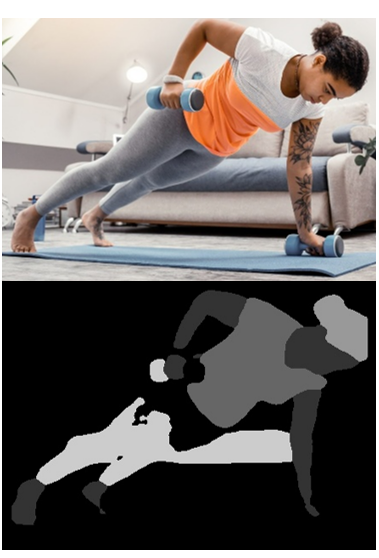}}
    \caption{ (a) Self-collision on body mesh (red circles). (b) A fitting example with inaccurate recovery of dress and pose due to the absence of dress in the training data and depth ambiguity. (c) Noisy segmentation result of \cite{yang20}.}
    \label{fig:limit}
\end{figure}